\documentclass[sigconf]{acmart}
\usepackage{algorithm}
\usepackage{algpseudocode}
\newcommand{\Input}{\item[\textbf{Input:}]}
\newcommand{\Output}{\item[\textbf{Output:}]}

\AtBeginDocument{%
  }

\setcopyright{acmlicensed}
\copyrightyear{2018}
\acmYear{2018}
\acmDOI{XXXXXXX.XXXXXXX}
\acmConference[Conference acronym 'XX]{Make sure to enter the correct
  conference title from your rights confirmation email}{June 03--05,
  2018}{Woodstock, NY}
\acmISBN{978-1-4503-XXXX-X/2018/06}

\begin{document}

\title{Skewness-Guided Pruning of Multimodal Swin Transformers for Federated Skin Lesion Classification on Edge Devices}

\author{Kuniko Paxton}
\email{k.azuma-2021@hull.ac.uk}
\orcid{1234-5678-9012}
\authornotemark[1]
\affiliation{%
  \institution{University of Hull}
  \city{Hull}
  \country{United Kingdom}
}

\author{Koorosh Aslansefat}
\affiliation{%
 \institution{University of Hull}
 \city{Hull}
 \country{United Kingdom}}
\email{K.Aslansefat@hull.ac.uk}

\author{Dhavalkumar Thakker}
\affiliation{%
  \institution{University of Hull}
  \city{Hull}
  \country{United Kingdom}}
\email{D.Thakker@hull.ac.uk}

\author{Yiannis Papadopoulos}
\affiliation{%
  \institution{University of Hull}
  \city{Hull}
  \country{United Kingdom}}
\email{Y.I.Papadopoulos@hull.ac.uk}


\begin{abstract}
In recent years, high-performance computer vision models have achieved remarkable success in medical imaging, with some skin lesion classification systems even surpassing dermatology specialists in diagnostic accuracy. However, such models are computationally intensive and large in size, making them unsuitable for deployment on edge devices. In addition, strict privacy constraints hinder centralized data management, motivating the adoption of Federated Learning (FL).
To address these challenges, this study proposes a skewness-guided pruning method that selectively prunes the Multi-Head Self-Attention and Multi-Layer Perceptron layers of a multimodal Swin Transformer based on the statistical skewness of their output distributions. The proposed method was validated in a horizontal FL environment and shown to maintain performance while substantially reducing model complexity. Experiments on the compact Swin Transformer demonstrate approximately 36\% model size reduction with no loss in accuracy. These findings highlight the feasibility of achieving efficient model compression and privacy-preserving distributed learning for multimodal medical AI on edge devices.

\end{abstract}



\received{30 September 2025}

\maketitle

\section{Introduction}
In recent years, the application of advanced computer vision has expanded into diverse fields with no end in sight, and its adoption is progressing within the medical sector. Among these models, skin lesion classification models, for instance, have been developed with performance surpassing that of dermatologists \cite{esteva2017dermatologist}. The models underpinning this high accuracy, such as Convolutional Neural Networks (CNNs) and Vision Transformer \cite{dosovitskiy2020image} with their backbones pretrained using a large image dataset, ImageNet \cite{deng2009imagenet}, tend to be computationally expensive and require a large capacity \cite{liu2020pruning}. This burden increases significantly as model structures become deeper and larger. Given that skin lesion classification models are frequently deployed in scenarios with limited computing resources, such as being embedded in Mobile Health Applications or used in medical Edge AI devices for use in clinics, there is a pressing need to reduce their computational weight.
Additionally, in the medical domain, collecting individual patient health metadata \cite{ukgovernment2025,dataprotectionact2018} is challenging for conventional centralized models due to data protection laws and stringent data privacy requirements. Moreover, aggregating extensive image data to central servers requires a significant amount of communication cost between clients and servers. Against this background, distributed learning approaches, such as Federated Learning (FL) \cite{mcmahan2017communication}, are gaining prominence. Indeed, in the case of FL, only parameters and biases are generally aggregated to the server. However, the number of parameters of the smallest size of Vision Transformer exceeds 80 million. To reduce the server up and down streaming cost, there is a great demand for reducing the number of parameters and downsizing the model.
As a further new trend, we need to note that it is increasingly recognised that these skin lesion classification models—which previously relied solely on images for condition classification—can achieve even higher diagnostic performance \cite{vachmanus2023deepmetaforge, de2022exploring}. This is accomplished by effectively incorporating health metadata as input and enabling the models to learn in a multimodal manner.
In other words, the compact model, which is compatible with Edge devices, can be applied in FL environment. In addition, compact models must also support multimodal input rather than handling single inputs only. To address these challenges, we propose a novel pruning method, which guides the multimodal model's focus area into the skin lesion site without complicated analysis. Inspired by \cite{paxton2025enhancing}'s centralized, unimodal ViT approach on early layers, we instead structurally prune Multi-Head Self-Attention (MSA) and Multi-Layer Perceptron(MLP) units across encoder blocks in a multimodal Swin Transformer \cite{liu2021swin} under FL. Various model compression techniques have been proposed in prior research. Combining the compression techniques is also popular, but we selected pruning as it is a foundational method to compress due to minimizing model structure before applying other methods and its compatibility with other methods. Our core approach lies in the characteristics of the dataset. Images with a limited purpose, such as dermatological lesion images, lack the colorful and complex composition found in general-purpose datasets like ImageNet or CIFAR-10 \cite{krizhevsky2009learning}. Consequently, we can statistically identify the regions the model should focus on in broad terms. This manuscript, therefore, focused on the computationally intensive MSA and MLP layers within the Transformer architecture. By measuring the statistical skewness of their respective output distributions, components generating outputs not focused on skin lesions were pruned. Moreover, we introduce a framework to continuously calibrate unnecessary model structure via pruning on the server side in the horizontal FL, while maintaining its performance.

\section{Related Work}
This manuscript adopted the Swin Transformer \cite{liu2021swin} as the image encoder, which is one of the Vision Transformers, as the target model for pruning. This is because research aimed at improving the accuracy of skin lesion classifiers has demonstrated that transformers exhibit higher performance than CNNs, and they have become the mainstream model since 2020. For these reasons, this section focuses on pruning techniques targeting transformers.
\subsection{Structured Pruning Method in Transformers}
Pruning methods for transformers can be categorized into four types. The first focuses on similarities within the model structure, removing redundant tokens and filters \cite{rao2021dynamicvit}. Research has introduced the Wasserstein distance as a similarity metric between tokens \cite{sun2022filter}. These approaches are based on structural redundancy and, like ours, do not rely on semantic features. The second method prunes patches and multi-head attention based on importance calculated using saliency \cite{naseer2021intriguing}, entropy \cite{lee2018snip}, and the L2 norm \cite{pachon2023efficient}. There are also studies that prune MLP layers using the L1 norm \cite{zhu2021vision}, but none of them consider semantics. The third approach is a strategy that uses explainability methods, such as those employed by \cite{dutta2024vtrans} and \cite{yu2023x}, to retain structures that are important for prediction. However, these are complex methods, in contrast to our approach, which is based on simple statistics, such as skewness. The fourth approach measures importance at the encoder block level, rather than the layer level \cite{zheng2022savit,yang2023global,chen2021chasing}. Whilst our method also considers block-level relationships, it achieves this through block-wise training. Furthermore, whilst existing research assumed highly redundant Vision Transformers, our approach differs significantly by further downsizing the already efficiency-optimized Swin Transformer.
\subsection{Pruning in Federated Learning Setting}
\cite{jiang2023complement} proposed Complement Sparsification. It reduces communication cost by masking out unnecessary structure to zero during each round of server aggregation. However, this method does not prune the model size itself, which is retained by clients or servers. In contrast, our approach differs in that it can reduce the size of the model structure itself. \cite{fan2024data} proposed a method that only averages overlapping parameters with weights when clients have different devices. \cite{gao2024device} adopted a strategy of performing selective pruning as an optimization problem during distribution from the server to the client. It is specialized for CNNs and cannot be applied to different architectures, such as transformers. \cite{huang2023distributed} proposed a method where only pruned clients with low mean loss for batch normalization are aggregated to the server. This constitutes an 'aggregation strategy' rather than a pruning technique itself. The pruning method was top-k of the parameter importance. FedMEF \cite{huang2024fedmef}, and \cite{jiang2022model,jiang2023computation} employed pruning tailored to client resource limitations, and \cite{li2024model} specialized in the agricultural domain, with each client performing individually optimized pruning. All these approaches differ from ours in that clients perform pruning. Meanwhile, \cite{wu2023efficient} stated that client-driven pruning carries significant inconsistency risks and argued that server-led pruning is well-suited for FL. FedPE \cite{yi2024fedpe} automatically explores the optimal subnet for each client in every round, repeatedly pruning and expanding based on changes in performance and accuracy. Additionally, FL-PQSU \cite{xu2021accelerating} is a pipeline that combines structural pruning, weight quantization, and selective updates. Their experiments used only CNN-based models. Lastly, an approach by \cite{lin2022federated} also performs pruning on the server side, but both rely on importance scores based on the absolute value of weights.

\subsection{Research Questions}
From the aforementioned background, the following research questions are derived:
\begin{itemize}
\item \textbf{RQ1:} Can the pruning based on the skewness of attention activations reduce the multimodal model size to suit edge devices while maintaining their performance?
\item \textbf{RQ2:} Is it possible to create a framework whereby multimodal models can be reduced in size through pruning calibration during server aggregation, using the same method for a single model?
\item \textbf{RQ3:} Can the RQ2 framework maintain accuracy while reducing its size through repeated iterations within a horizontal FL environment?
\end{itemize}

\subsection{Main Contributions}
Our main contribution by responding to our research questions is as follows:
(1) We proposed a new method for structural pruning of the Swin Transformer using the skewness of the score distribution of the attention activations as a metric for all blocks and heads.
(2) Our approach has demonstrated that model compression to compatible edge devices while maintaining high classification accuracy.
(3) We confirmed that our proposed pruning method is effective with multimodal learning that integrates medical images and metadata, demonstrating its versatility beyond a single model.
(4) We demonstrated that our skew pruning framework can maintain high performance in an FL setting.

\section{Methodology}
\begin{figure*}
    \centering
    \includegraphics[width=1\linewidth]{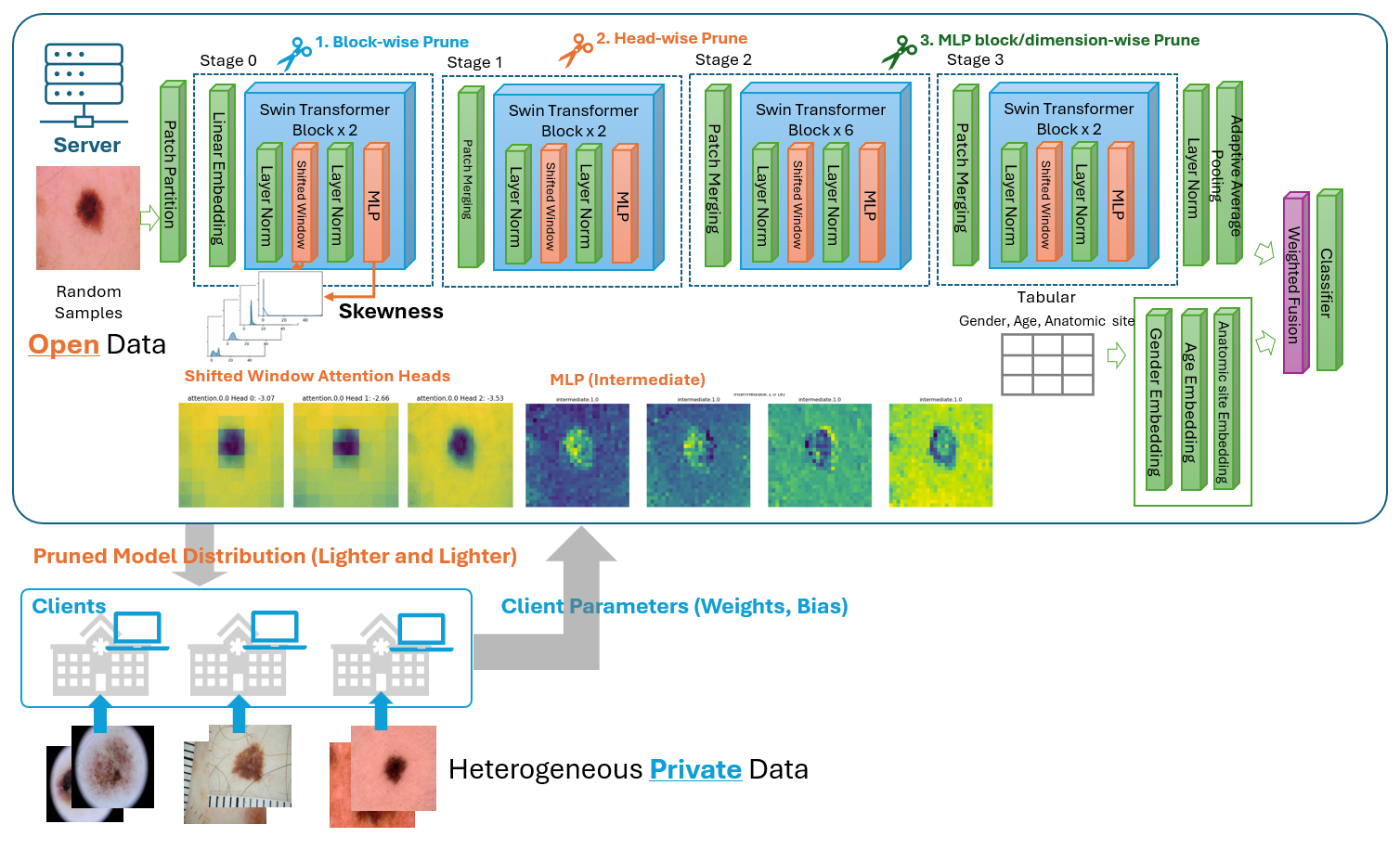}
    \caption{Skewness Pruning and Calibration in Horizontal FL Framework}
    \label{fig:framework}
\end{figure*}

Figure \ref{fig:framework} illustrates our pruning methodology and its overall execution within the FL environment. Pruning is always performed on the server side, utilising open data that requires no consideration for privacy. On the server side, Targets are the MSA and MLP within the encoder block of the image feature extraction. Rather than employing different methods as in existing research, both MSA and MLP are pruned using the same technique.. Subsequently, the reduced model is distributed to the client, where it trains using diverse and unknown data. Finally, parameters are aggregated on the server. Pruning applies to any redundant structure that remains at this stage, thereby establishing a framework that consistently maintains a compact model.

\subsection{Skewness Pruning Technique}
This subsection describes the process of selecting structures for pruning based on the projected output of the MSA head and the linear layer of the MLP.
\subsubsection{Attention Head in MSA}
\label{se:attention_head}
Given MSA attention output (projected values) $A \in \mathbb{R}^{B\times W^{2}\times H \times D}$, where $B$ is the batch size, $W$ is the shifted global window size, $H$ is the number of heads, and $D$ is the head dimension. The aggregated empirical distribution is defined,
\begin{equation}
    \Pi = \left\{ \left\| a^{(0)}_{i,j,h} \right\|_{2} \mid (i,j), h \right\}
\end{equation}
for the first batch $A_{0} \subset A$. We vectorize to $V = \text{vec}(\Pi)\in \mathbb{R}^{W^{2}}$. For each head $h$, the skewness is computed. Given the mean
\begin{equation}
\label{eq:mean}
    \mu_h = \frac{1}{W^2} \sum_{i=1}^W \sum_{j=1}^W v_{h,(i,j)},
\end{equation}
Then, the skewness $s_{h}$ each head is,
\begin{equation}
\label{eq:skew}
    s_{h} = \frac{\frac{1}{W^{2}}\sum_{i=1}^{W}\sum_{j=1}^{W}(v_{h,i,j}-\mu_{h})^{3}}{(\frac{1}{W^{2}}\sum_{i=1}^{W}\sum_{j=1}^{W}(v_{h,i,j}-\mu_{h})^{2})^{\frac{3}{2}}}
\end{equation}

When skewness is $s_{h} > 0$, the head is interpreted as having a focus on skin lesions. In contrast, when skewness is $s_{h} \le  0$, it is deemed not to be focusing on beneficial information and is pruned.
\subsubsection{Intermediate Linear Layer in MLP}
\label{se:intermediate}
Given intermediate output 
\begin{equation}
    Z\in \mathbb{R}^{B\times W^{2}\times C}, C = r \cdot  d
\end{equation}
Where $r$ is a ratio to expand a feature vector expression, and $d$ is the hidden embedding dimension. The group is created $G$ alternative of heads in MSA by dividing $C$ by $r$. For the first batch $z^{(0)}\subset Z$, Similar to the case of MSA, 
\begin{equation}
    \Pi = \left\{ \left\| z^{(0)}_{i,j,g} \right\|_{2} \mid (i,j), g \right\}
\end{equation}
Then vectorize $\Pi$. Afterwards, the skewness is calculated using Eq. \ref{eq:mean} and \ref{eq:skew} as the same as MSA.

\subsection{Structured Pruning}
\label{se:prune}
Firstly, we explain the case where all attention heads in a single encoder block of Shifted Window MSA are subject to pruning. The MSA is normally as follows (refer to \cite{vaswani2017attention}):
\begin{equation}
    \text{MSA Head}(Q,K,V) = \text{Concat}(\text{head}_{1},.., \text{head}_{h})W^{o},
\end{equation}
where 
\begin{equation}
    \text{head}_{i}=\text{Attention}(QW_{i}^{q}, KW_{i}^{k}, VW_{i}^{v}).
\end{equation}
Pruning all heads removes the MSA output. However, to maintain residual connections and stable training, we approximate it with the identity map Eq. \ref{eq:identity_map}.
\begin{equation}
\label{eq:identity_map}
    \text{MSA Head}(Q,K,V)\approx I(Q)=Q
\end{equation}
When not all heads need to be pruned, we denote $H_{\text{head}}$ as the set of heads to keep.
\begin{equation}
    \text{MSA Head}_{\text{keep}}(Q,K,V) = \text{Concat}(\left\{ \text{head}_{i}|i\in H_{\text{keep}} \right\})
\end{equation}
and for bias, $B_{\text{keep}} = B\left[ :, H_{\text{keep}} \right]$.

Next, it is when all the Intermediate layers of an MLP block. Notations refer to \cite{liu2021swin}. MLP layers are defined
\begin{equation}
    z^{l+1}=\text{MLP}(LN(\hat{z}^{l+1}))+\hat{z}^{l+1}
\end{equation}
where $\hat{z}^{l+1}$ is output of Shifted Window MSA. In our approach, since we prune the MLP part with an identity mapping, only the LayerNorm from the output of MSA remains in the residual connection (Eq. \ref{eq:mlp_full_prune}).
\begin{equation}
\label{eq:mlp_full_prune}
    z^{l+1}=LN(\hat{z}^{l+1})+\hat{z}^{l+1}
\end{equation}
Lastly, we describe that case, partially indexing the intermediate to select for pruning. The MLP module of the Swin Transformer consists of LayerNorm, two linear transformations, and a GeLU activation in between them (Eq. \ref{eq:mlp}).
\begin{equation}
\label{eq:mlp}
    MLP(\hat{z}^{l+1})=\text{GeLU}(\hat{z}^{l+1}W_{1}+b_{1})W_{2} + b_{2}
\end{equation}
$W_{1}$, $b_{1}$ are weights and bias the first linear transformation and $W_{2}$, $b_{2}$ are the second ones. Given indices to keep in intermediate $\text{Indices}_{\text{keep}}$, pruned the first linear transformation becomes $\acute{W_{1}}=\text{Indices}_{\text{keep}}^{-1}W_{1}$ and $\acute{B_{1}}=\text{Indices}_{\text{keep}}^{-1}B_{1}$. After GeLU, the input dimension of the second linear transformation is pruned $\acute{W_{2}}=\text{Indices}_{\text{keep}}W_{2}$ accordingly. In the end, pruning is expressed with Eq. \ref{eq:mlp_prune}
\begin{equation}
\label{eq:mlp_prune}
    MLP(\hat{z}^{l+1})=\text{GeLU}(\hat{z}^{l+1}\acute{W_{1}}+\acute{b_{1}})\acute{W_{2}} + b_{2}
\end{equation}

\subsection{Pruning Application in Fine-Tuning Process}
The procedure using the equations in the above subsection constitutes the Skew Prune algorithm \ref{al:finetunrn}.
\begin{algorithm}
\caption {Stepping Pruning of Image Encoders}
\label{al:finetunrn}
\begin{algorithmic}[0] 
\Input Pretrained Swin Transformer $M$
\Output Pruned Model $M$
\State Load model $M\acute{}$
\State Fine-Tuned model $M$
\For {Swin Stage $s$ = 0 to $\text{stage count}(s) - 1$ }
\For {Block $b$ = 0 to $\text{block count}(s) - 1$ }
\State Select attention head $h$ in $b$ in $s$ to be pruned computed by the strategy in subsection \ref{se:attention_head} from $M$
\State Prune $h$ using the equations subsection \ref{se:prune}
\State Select intermediate $g$ in $b$ in $s$ to be pruned computed by the strategy in subsection \ref{se:intermediate} from $M$
\State Prune indices of $g$ using the equations subsection \ref{se:prune}
\EndFor
\State Freeze parameters of $s$
\State Fine-Turned model $M$ to $M\acute{}$
\EndFor
\State Return $M\acute{}$
\end{algorithmic}
\end{algorithm}

\subsection{Evaluation Metrics}

\begin{itemize}
    \item \textbf{Performance}: Pruned model performance is evaluated using accuracy and F1 score. In the FL setting, model performance is evaluated using the server model, as it corresponds to the final global model.
    \item \textbf{Model Running Cost}: The reduction in the structural size of the model is measured by the total number of model parameters, the saving in the amount of computation cost required per process during prediction is measured by Floating Point Operations per Second (FLOPS), and the decrease in the amount of memory required during execution is evaluated by Memory Footprint (MF).
    \item \textbf{FL Operational Cost}: The operational cost of FL is mainly determined by the file size of the model distributed to client devices. This is because, when the number of communication rounds between the server and client is fixed, the difference in the final communication cost depends on the difference in model size due to the two-way communication (upload and download) between the client and server.
\end{itemize}

\section{Experimental Setup}
\textbf{Dataset:} In this study, Human Against Machine with 10000 training images (HAM) \cite{tschandl2018ham10000,codella2019skin} dataset is selected, which is publicly available and widely used in skin lesion classification research. This dataset comprises seven classifications: Actinic Keratoses, Benign Keratoses, Basal Cell Carcinoma, Dermatofibroma, Melanocytic Nevi, Melanoma, and Vascular. In the FL environment, 20\% of all data was first set aside as test data on the server side. Subsequently, the remaining data was split into training and validation data according to the number of clients. Each client's data comprises 80\% training data and 20\% validation data.
\textbf{Multimodal Model:} In the multimodal model, the Swin Transformer was employed for image feature extraction, while tabular data (gender, age, lesion location) were each embedded using lookup embedding. These were added together just before the classifier, with weights of 85\%, 5\%, 5\%, and 5\% assigned to the image and each feature, respectively. This ratio was confirmed as optimal through our preliminary experiments.
\textbf{FL settings:} FL employed the Flower library \cite{beutel2020flower}, with emulation experiments conducted using an NVIDIA GeForce RTX 4090 (1 GPU) and 24 CPUs. Six clients participated, with training conducted over 100 rounds, each round comprising 3 epochs. Training was performed to obtain the initial pruned model, as this setting provided the best performance in F1 score during experiments.

\section{Result}
\subsection{Effectiveness of Skewness Structured Pruning}
\label{se:result_skewness}
The result summary is shown in Table \ref{tb:prune}. The baseline model before pruning demonstrated good performance with an accuracy of 85.1\% and an F1 score of 76.9\%, but it had a high computational cost of 4.36 GFLOPS per inference. Its parameter number of 27.7 million was exceptionally high for a Tiny model, and both its memory footprint and model file size exceeded 100MB. The pruned model based on skewness maintained performance with less than 1\% accuracy degradation and only about 2\% deterioration in F1 score. Meanwhile, computational cost per inference was reduced to 2.32 GFLOPs, approximately 53\% of the pre-pruning value. At the same time, parameter count, memory footprint, and model size were all scaled down to about 38\% of their pre-pruning levels.

Next discussion is the pruning results in the FL environment. The model performance prior to pruning was slightly lower compared to that of the single multimodal model. This stems from the adoption of simple averaging for aggregation during client aggregation, which clearly ascertains the effect of pruning. Nevertheless, under privacy protection constraints that prevent centralized data aggregation, accuracy remained at approximately 84\%, and the F1 score was around 70\%. The result of pruning the model after server aggregation is referred to as 'FL Pruned'. A little decrease in accuracy was observed. Furthermore, advancing pruning to Stage 3 reduced the inference cost by approximately half, while the parameter count, memory footprint, and model size were each reduced to approximately 36\% of their original values. These results confirm that the proposed pruning method achieves equivalent or superior model compression effects not only in single-model scenarios but also within FL environments.

\begin{table*}[]
\centering
\begin{tabular}{l|c|c|c|c|c|c}
\hline\hline
                      & Single & Pruned & Effects & FL & FL Pruned & Effects \\ \hline
Accuracy              & 0.851    & 0.844& 0.65\% & 0.838 & 0.838 & 0 \% \\
F1                    & 0.769    & 0.747& 2.14\% & 0.709 & 0.696 & 0.01\% \\
GFLOPS                & 4.36     & 2.32& 53.21\% & 4.36 & 2.16 & 49.54\% \\
Parameters (M)        & 27.70     & 10.44& 37.69\% & 27.70 & 9.99 & 36.06\% \\
Memory Footprint (MB) & 106.02   & 40.14& 37.86\% & 106.02 & 38.45  & 36.27\% \\
Model Size (MB) & 108.67   & 41.19& 37.90\% & 108.6 & 39.45 & 36.33\% \\ \hline\hline
\end{tabular}
\caption{Evaluation Result}
\label{tb:prune}
\end{table*}

\section{Discussion}
\subsection{Discussion}
\textbf{RQ1:} As shown in the result section \ref{se:result_skewness}, the pruning strategy proposed in this study (a method that prunes networks not to focus on lesion regions by targeting computationally intensive attention heads and MLP intermediate layers using skewness of distribution of output) enabled significant reductions in processing cost and model size while maintaining accuracy. These results demonstrated the effectiveness of our approach. \textbf{RQ2:} In this study, we applied the pruning technique, which is effective for a single model, to models after server aggregation in the FL environment. The results showed achievement of compression rates comparable to those for a single model. This suggests that the proposed method functions stably even when data distributions differ across clients. Consequently, we confirmed that the same pruning algorithm used for a single model can be applied to the FL environment. \textbf{RQ3:} Repeatedly performing pruning and training across multiple rounds within the FL environment confirmed that accuracy could be maintained stably while sustaining reductions in inference cost and model size. Notably, even when pruning was advanced to stage 3, the degradation in accuracy was minimal. Our proposed method enables pruning while maintaining prediction accuracy even under FL.

\subsection{Research Limitations and Future Work}
This study aimed to reduce model size under diverse conditions, both single models and horizontal FL, by pruning different model structures (MSA and MLP) consistently based on the skewness of the output distribution. Consequently, the multimodal fusion and FL aggregation methods were deliberately kept simple to enable clear evaluation of the pruning's effectiveness. Moving forward, we plan to explore advanced FL aggregation techniques and multimodal fusion technologies along with low-resolution learning. This will be done with the objectives of supporting smaller edge devices and reducing communication costs.

\section{Conclusion}
This study, unlike previous research, employed the already compacted Swin Transformer as its backbone and tackled multimodal skin lesion classification using image and limited feature tabular data. It demonstrated that a single pruning technique can be applied to different model blocks. Furthermore, this approach was confirmed to deliver equivalent effectiveness not only in a single model but also within a Horizontal FL environment. This research demonstrates the potential for downsizing and practical implementation of multimodal models in environments where edge devices are becoming increasingly prevalent.


\bibliographystyle{ACM-Reference-Format}
\bibliography{main}

\appendix

\end{document}